# A Machine Learning Approach to Predicting Single Event Upsets

Archit Gupta, Chong Yock Eng, Deon Lim Meng Wee, Rashna Analia Ahmed, See Min Sim
Nanyang Technological University, Singapore
{archit001, ceng006, dlim074, rashna002, ssim033}@e.ntu.edu.sg

*Abstract*—A single event upset (SEU) is a critical soft error that occurs in semiconductor devices on exposure to ionising particles from space environments. SEUs cause bit flips in the memory component of semiconductors. This creates a multitude of safety hazards as stored information becomes less reliable. Currently, SEUs are only detected several hours after their occurrence. CREMER—the model presented in this paper—predicts SEUs in advance using machine learning. CREMER uses only positional data to predict SEU occurrence—making it robust, inexpensive and scalable. Upon implementation, the improved reliability of memory devices will create a digitally safer environment onboard space vehicles.

## I. INTRODUCTION

Many space bodies like satellites and the International Space Station (ISS) lie in the Low Earth Orbit (LEO), resulting in great exposure to cosmic microwave background radiation, solar radiation from the sun and Van Allen radiation. Although most radiation gets deflected by the Earth's magnetosphere, high energy charged particles may still pass through and disrupt the operation of space devices. These charged particles, when passing through a medium, lose energy by ionisation and induce electron-hole pairs along their path. The interaction of electrons, protons, and heavy ions with integrated circuits could therefore lead to a total dose degradation, or worse, a single event effect—a class of radiation effects in electronic devices. In this research paper, we will be focusing on predicting single event upsets (SEU)—a subset of single event effects.

In digital memory and logic devices, SEUs are non-destructive 'soft' errors. They normally appear as a bit flip in memory units or transient pulses in logic devices, and do not permanently destruct the functions of a device. Therefore, error detection and correction codes are frequently adopted to reduce the impact of SEUs. However, modern semiconductor devices tend to have tiny junction areas with proportionately small amounts of charge to control the state of memory units. This increases SEU disturbances because a single heavily charged particle passing through the junction would be sufficient to induce charge into the node and change its state, causing disruption of data stored at the node. As modern semiconductor devices are getting even smaller, predicting and mitigating SEUs is becoming an increasingly important problem. Our research is focused on mitigating the impact of single event upsets by employing machine learning. The model presented will allow the prediction of SEUs in memory units—prompting timely mitigation of the occurrence.

Previous studies have been done on a similar problem [1] [2], however, our approach differs by utilising a novel method on a more constrained dataset.

## II. BACKGROUND

Current models used to predict SEUs are largely based on the transfer of energy from protons to electrical components as they pass through Very Large-Scale Integration (VLSI) devices [3]. These models are mostly tested in controlled environments with the aim of predicting critical energy levels of VLSI devices and to verify the strength of the shielding methods in place. The calculations behind these models are based on data collected in labs under the assumption that a device's operation in such simulated environments is an indication of its operation in space. As a result of such assumptions, traditional physics-based models are not foolproof or robust.

*Linear Energy Transfer (LET)* is the fundamental idea behind several SEU prediction models. LET is the deposition of energy by high energy protons onto the cells of a semiconductor device as it pierces through the device. The amount of energy transferred from the proton depends on the incident angle and the route it takes through the device. This energy transfer is described by the following equation:

$$L_{eff} = \frac{L}{cos\theta}$$

where *L*- is the net energy transferred to a cell if the path of the proton is perpendicular to the cell, and $\theta$ is the angle of entry into the cell. If the charged energy deposited from the proton into the cell exceeds a certain amount (critical charge), an SEU can occur.

*Path-Length Distribution Model* is a model that is currently used to test the likelihood of an SEU occurring in semiconductor devices. It uses variables such as the differential distribution of path lengths to monitor the charge with respect to the critical value for a device. SEUs occur when the charge exceeds a device's critical value. However, a drawback is that the sensitive volume is assumed to be that of a rectangular parallelepiped across each cell in the device. In reality, cells are dynamic and the volumes of individual cells cannot be generalised. For example, there may be multiple upsettable nodes with differing sensitive volumes for each microprocessor.

*Bendel Proton Upset Model* is another model that is currently used to provide an estimate of the SEU occurrence rate in a semiconductor device using semi-empirical fit parameters [4]. In particular, it considers high energy protons that produce secondary particles with energy high enough to cause an SEU. Since the SEU occurrence rate is calculated based on the mean value in a hyper dynamic environment, the short term rate of SEU occurrence may be much higher than the estimate. Therefore, systems with semiconductor devices susceptible to SEUs need to have built in shielding to withstand short periods of greatly increased SEU occurrence rates. This takes away the value of predicting an SEU.

### III. METHODOLOGY

#### A. DATA

To build a model, we first needed to obtain data. Given the niche nature of our research, relevant data was not readily available. The data we used in training CREMER was obtained from the Institute of Space Systems, University of Stuttgart (Germany) [5]. The dataset contained the orbital positions, altitudes and timestamps of the SEU observed in an onboard memory device of the Low Earth Orbit "Flying Laptop" satellite mission during its in-orbit operation.

The dataset only accounted for positive cases—the data for when an SEU had occurred. Thus, in order to create a realistic dataset, it was critical to generate data for when an SEU had not occurred—the negative cases. The methodology of generating data is as follows. The first and last dated SEU occurrence within the provided data was 2017-08-15 (01:24:30) and 2018-05-28 (5:46:25) respectively. The elapsed time was 412,101 minutes (rounded down). Assuming that the scrubbing rate aboard the mission was strictly followed, a total of 206,050 (rounded down) scrubbings were executed. The dataset contained 2130 data points—data for when an SEU was recorded. Hence, it is reasonable to deduce that there were approximately 203,920 scrubbings conducted in which an SEU was not detected.

A random altitude was generated from a Gaussian distribution. The longitude and latitude were generated randomly over a range based on the minimum and maximum of their respective data columns. Random uniform sampling of a point inside a unit square can exhibit clustering of points, as well as regions that contain no points at all. Hence, it is reasonable to consider a Poisson disk sampling method in order to generate the desired distribution.

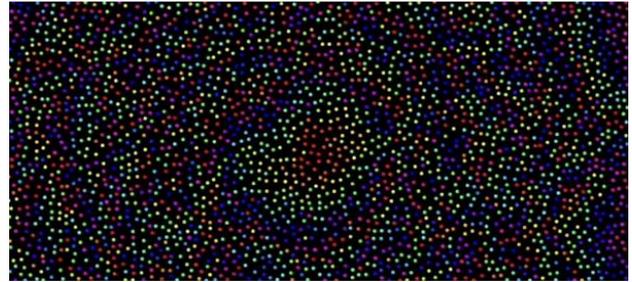
Figure 1: Poisson Disk Sampling Distribution

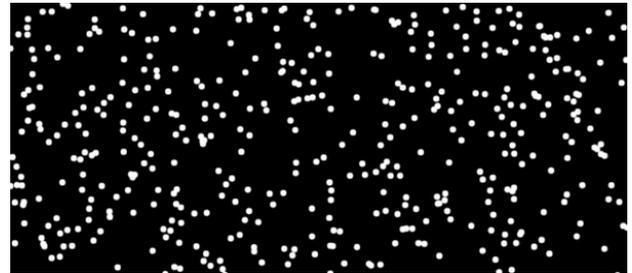
Figure 2: Random Uniform Sampling Distribution

However, before we could train classification models, we needed to address a few issues. Firstly, the dataset was highly imbalanced, with a class ratio of 1:94. Secondly, as we are building a real-time model that will train periodically, training time has to be as small as possible to minimise downtime. To account for both of these issues, we performed undersampling on the training data by generating centroids based on clustering methods. This not only produced a balanced dataset for training purposes but also reduced our training size drastically, allowing for a much faster model. The distribution of our testing data remained unchanged.

#### B. MODEL

CREMER uses a soft voting ensemble to combine extreme gradient boosting and random forests. The main purpose of this was to employ both boosting and bagging to make this model as robust as possible in the face of uncertain and unique space environments. As we were building CREMER to work on different datasets collected under different space environments, hyperparameter tuning techniques like grid search and random search were not viable.

Instead, we resorted to manually tuning certain hyperparameters to make both the classification algorithms perform well on imbalanced datasets without overfitting. A technique common to both algorithms was scaling the gradient for the positive class by assigning it a higher weight. This was done using the *class_weight* and *scale_pos_weight* parameters in random forest and extreme gradient boosting classifiers respectively. The value was chosen by approximating the ratio between the positive and negative class samples. While building this model, we chose recall as the metric to optimise for. Essentially, we wanted CREMER to correctly classify as many SEUs as possible, even if some non-SEUs were misclassified as SEUs. This trade-off is reasonable under our assumption that the SEU mitigation techniques implemented are non-invasive and inexpensive.

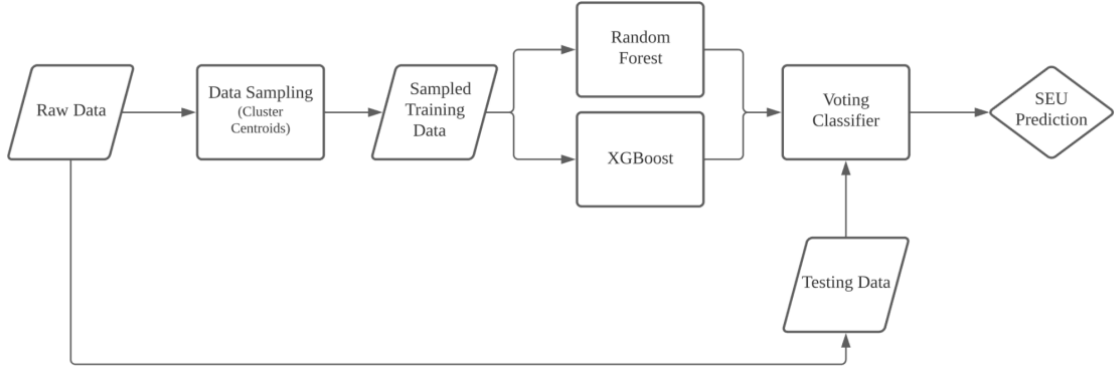

Figure 3: Model Training Pipeline

The pseudocode for our algorithm is as follows

```
Algorithm 1 CREMER
  procedure DATASAMPLING(x_train, y_train)
    x_train, y_train = ClusterCentroids(voting=soft).fit_resample(x_train, y_train)
    Return x_train, y_train
  end procedure

  procedure MODELBUILDING
    classifier_rf = RandomForestClassifier()
    classifier_xgb = XGBClassifier()
    classifier_voting = VotingClassifier(estimators=[classifier_rf, classifier_xgb], voting=soft)
    Return classifier_voting
  end procedure

  procedure MODELTRAINING(x_train, y_train)
    x_train, y_train = DataSampling(x_train, y_train)
    classifier = ModelBuilding()
    classifier.fit(x_train, y_train)
    Return classifier
  end procedure
```

Figure 4: Algorithm Pseudocode

## IV. RESULTS

Following standard practice, we split our dataset—80% is used for training and 20% is used for testing. All the results presented in this paper were obtained by running CREMER on the test dataset. We resampled and retrained our algorithm 100 times—achieving an average recall of 0.972 and an average precision of 0.015. On average, CREMER takes 79.44 seconds to train, but only 0.007 milliseconds to make a prediction.

Upon tuning certain hyperparameters, like the positive class weight, we achieved a higher precision and AUROC (Area Under the Receiver Operating Characteristic) scores at the expense of recall. The values obtained below were averaged across 10 runs. In all of these cases, the weight of the negative class was fixed at 1. CREMER has high economic feasibility given the ease of implementation of the model.

| Positive Class Weight | Recall | Precision | AUROC |
|---|---|---|---|
| 100 | 0.967 | 0.015 | 0.635 |
| 50 | 0.963 | 0.015 | 0.629 |
| 10 | 0.930 | 0.016 | 0.646 |
| 5 | 0.925 | 0.016 | 0.660 |
| 1 | 0.878 | 0.019 | 0.692 |
| 0.1 | 0.755 | 0.021 | 0.686 |

Table 1: Impact of Class Weight on Performance Metrics

## V. FEASIBILITY ANALYSIS

As CREMER uses traditional classification to train on real-time data, it avoids the high computational and R&D costs of traditional physics-based models. Although there is ongoing research on SEU estimation techniques such as estimating event rates from cross-section data, their costs for execution are high. Moreover, such new models would require additional time and resources to be redesigned for different space vehicles, a limitation the model presented doesn't possess.

Lastly, this model has a very fast turnaround time due to the prediction algorithm being fully online. This means that we can evaluate the results and make relevant updates to the algorithm periodically.

## VI. CONCLUSION & FURTHER WORK

In this paper, we presented CREMER—a flexible machine learning model that predicts SEUs under unpredictable space environments. This allows CREMER to be deployed under extremely constrained environments where traditional detection techniques and other ML models fail. Our current work presents the first model that can predict SEUs based solely on positional data.

Moreover, for larger space vehicles such as the ISS, CREMER can be made to automatically incorporate additional information such as tri-axial magnetic field, ambient pressure, particulate matter concentration, and $CO_2$ concentration data, in order to further improve its performance. Lastly, CREMER's Recall-AUROC trade-off can be adjusted using a single metric, making the model easily adjustable to space vehicles with varying functional requirements and SEU mitigation techniques.

## VII. REFERENCES & CITATIONS